\documentclass[letterpaper, 10 pt, conference]{ieeeconf}  

\IEEEoverridecommandlockouts                              

\usepackage{amsmath,amssymb,amsfonts}
\usepackage{graphicx}
\usepackage{textcomp}
\usepackage{xcolor}
\usepackage{todonotes}

\title{\LARGE \bf
Visualization of Intended Assistance for Acceptance of Shared Control
}

\author{Connor Brooks$^{1}$ and Daniel Szafir$^{2}$%
\thanks{$^{1}$University of Colorado Boulder, Department of Computer Science
        {\tt\small connor.brooks@colorado.edu}}%
\thanks{$^{2}$University of Colorado Boulder, Department of Computer Science and ATLAS Institute
        {\tt\small daniel.szafir@colorado.edu}}%
}

\begin{document}
\maketitle
\thispagestyle{empty}
\pagestyle{empty}

\begin{abstract}
In shared control, advances in autonomous robotics are applied to help empower a human user in operating a robotic system. While these systems have been shown to improve efficiency and operation success, users are not always accepting of the new control paradigm produced by working with an assistive controller. This mismatch between performance and acceptance can prevent users from taking advantage of the benefits of shared control systems for robotic operation. To address this mismatch, we develop multiple types of visualizations for improving both the legibility and perceived predictability of assistive controllers, then conduct a user study to evaluate the impact that these visualizations have on user acceptance of shared control systems. Our results demonstrate that shared control visualizations must be designed carefully to be effective, with users requiring visualizations that improve both legibility and predictability of the assistive controller in order to voluntarily relinquish control.
\end{abstract}

\section{Introduction}

Through \textit{shared control} \cite{abbink2018topology}, human intelligence can be combined with robot autonomy in order to aid direct human teleoperation. This control paradigm involves two separate controllers --- a human and an assistive controller --- working in concert to operate the same physical robot. While such systems are designed to employ advances in autonomous robotics in order to improve human-in-the-loop control, this requires the human and assistive controller to share a common goal. Any mismatch in goals or intended trajectories may result in these controllers working against one another. Thus, cooperation between these two controllers is critical for successful operation.

If a human's goal for a robotic system is known in advance, assistance can be provided without concern for goal misidentification \cite{philips2007adaptive, yu2003adaptive}. For more general applicability, the user's goal must first be inferred by an assistive system. Inferring a human's goal from their control inputs allows an assistive controller to estimate the target of assistance through the user's natural interactions with the system. Existing research has established various methods for performing such inference, ranging from proximity-based estimation \cite{goil2013using, katyal2014collaborative, kofman2005teleoperation} to the use of an Inverse Reinforcement Learning framework for reward function estimation \cite{dragan2012formalizing, javdani2015shared, schultz2017goal}.

\begin{figure}
\includegraphics[width=\columnwidth]{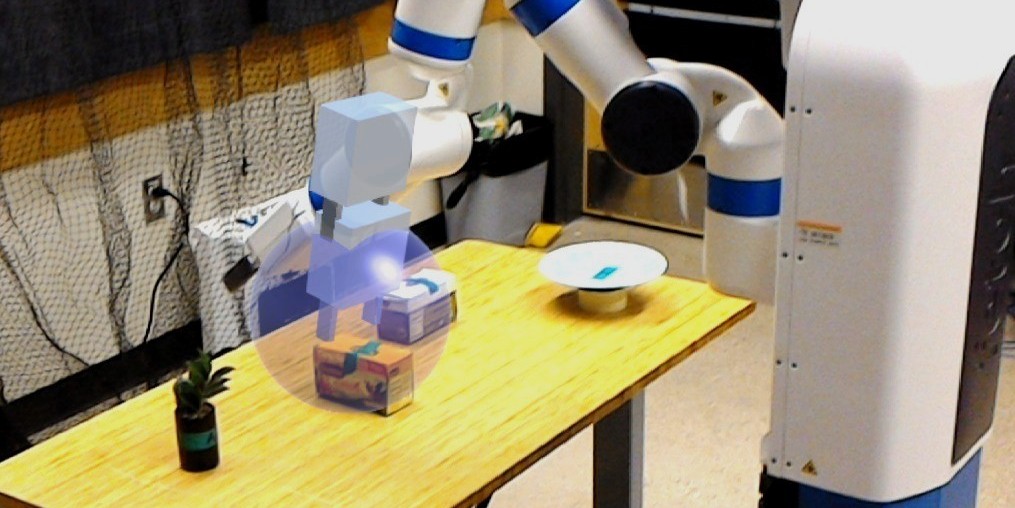}
\label{teaser_visualization}
\caption{Augmented Reality visualizations demonstrate the assistive controller's intended target and future trajectory keypoints.}
\end{figure}

Although shared control has been shown to improve efficiency, human teleoperators often display a hesitancy to use these systems \cite{dragan2013policy, brooks2019balanced}. We believe that one potential reason for this hesitancy is a lack of understanding of the assistive controller's intent. For example, an assistive controller might generate behavior that could be perceived as counter-productive (e.g. following a trajectory that temporarily moves away from the user's intended goal to overcome a joint limitation) and cause uncertainty for the user about whether the assistive controller has correctly inferred their goal.

In this work, we investigate how improving the communication between assistive controllers and users may improve user acceptance and adoption of shared control systems. In particular, we investigate how Augmented Reality (AR) can provide \textit{in situ} visualizations that directly communicate assistive controller goals and strategies to users during shared control of a robotic arm. Furthermore, we develop multiple levels of visualization in order to assess how much information is necessary to help users understand the behavior of an assistive controller. We evaluate these AR visualizations in a user study involving multiple types of assistive controllers in order to generalize this evaluation across different types of shared control systems. Finally, we present and discuss our results demonstrating that these visualizations make the shared control system more predictable to users and users are more willing to hand over control to the assistive controller when they are able to see its intended trajectory and goal.

\section{Related Work}

Despite assistive teleoperation systems improving user efficiency, prior work has found mixed results on human teleoperator preference for these systems versus those that give the user more control \cite{javdani2015shared, dragan2013policy, brooks2019balanced}. Dragan and Srinivasa \cite{dragan2013policy} suggest that, based on user feedback, it is important for the assistive controller to exhibit \textit{legible} behavior --- movement that helps the user understand its goal \cite{dragan2013legibility}. Our work follows directly from this idea through evaluation of two types of AR visualizations: one designed to make the assistive controller's goal legible and the other to make the robot's movement \textit{predictable} --- movement that matches the user's expectations.

Visualizations of future robot movement are notably useful in control systems that allow users to select from high-level robot behaviors \cite{tsui2011want}. Weisz et al. \cite{weisz2017assistive} use AR to display possible robotic arm grasps that a user iterates through and selects in order to issue commands. A similar design is behind the \textit{Director} control system designed by Marion et al. for use in the DARPA Robotics Challenge \cite{marion2017director} that uses visualizations of robot commands to help operators choose appropriate actions. In contrast to these systems, we are interested in assistive teleoperation involving continuous blending of direct user commands and assistive controller inputs throughout the task.

We use AR for displaying visualizations of assistive controller intent because this modality enables visualizations of information situated in the relevant real world locations. This situated information has been shown to be useful for human-robot interaction by improving human understanding of future robot movement \cite{walker2018communicating}, limiting context switching during teleoperation \cite{hedayati2018improving}, and conveying environmental information to human users \cite{yamamoto2012augmented} (for a further review of AR and human-robot interaction, see \cite{szafir2019mediating}). In particular, Rosen et al. \cite{rosen2020communicating} have shown that mixed-reality visualizations of trajectories enable users to better understand future robotic arm movement than 2D displays. Motivated by these strong results, we use AR visualizations to help users in understanding the assistive controller during shared control.

Our work is partially inspired by Zolotas et al. \cite{zolotas2018head}, who developed an AR system for shared control of an assistive wheelchair. Their system is intended to communicate the information that the assistive controller uses for planning in order to improve explainability. In further work, Zolotas and Demiris \cite{zolotas2019towards} evaluate the performance of users operating the wheelchair with and without their AR explanations and find that these explanations benefit user performance. However, the shared control systems investigated in these works do not contain goal inference, instead augmenting human control by handling complimentary tasks such as obstacle and collision avoidance. This results in shared control systems with a more indirect collaboration than those in which we are interested. Furthermore, our work extends this existing research by evaluating multiple levels of visualization across different types of assistive controllers. Finally, we investigate factors not addressed previously such as the impact of AR visualizations on the user's acceptance of shared control.

\textbf{Contributions: }
Our research addresses questions about shared control raised in prior work by evaluating the impact of visualizing an assistive controller's intent during shared control involving continuous collaboration between a human user and an assistive controller. Our contributions include an assessment of AR visualizations for assistive teleoperation of a robotic arm, the creation and assessment of multiple levels of visualization across different types of assistive controllers, and an analysis of how users handle control authority in response to AR visualizations.

\section{Hypotheses}

Informed by past research, we form three hypotheses about the impact of assistive controller intent visualizations on shared control:

\textbf{H1:} Visualizing assistive controller intent will increase user willingness to relinquish control in shared control.

\textbf{H2:} Visualizing assistive controller intent will improve user assessment of the shared control system's usability.

\textbf{H3:} Visualizing assistive controller intent will improve user assessment of the shared control system's predictability.

\section{System Design} \label{System_Description}
To test these hypotheses, we create an assistive controller for an object grasping task. Providing assistance in a cluttered environment requires inference of the user's goal, which is complicated by the possibility of different possible grasp positions for a given goal. We model the user's goal using a Hidden Markov Model (HMM) that allows us to take into account multiple grasps for a single goal.

\subsection{HMMs for Goal Inference}
Creating an HMM requires the determination of a set of possible hidden states $X$, the transition probabilities between each pair of these states $p(x'_{t+1} | x_{t})$, and the observation probabilities at each state $p(u_{t} | x_{t})$ given a set of possible observations $U$. Prior work in assistive teleoperation systems has utilized HMMs by letting the hidden state represent the goal of interest in the user's underlying intent while the observations at each state correspond to user commands \cite{aarno2005adaptive}. With this approach, the transition probabilities between states model the likelihood of changing goals during the task \cite{jain2018recursive}. 

We apply and extend this idea by creating separate states for each possible grasp of a goal object. Our hidden states can be grouped into classes corresponding to the enumeration of possible grasps for a particular goal. Thus, each hidden state, $x$, is a member of a single class $C_{\phi}$ corresponding to the goal $\phi$ for which that state represents a particular grasp. The probability of a goal given a sequence of observations $\xi_{0\rightarrow t} = (u_{0},...,u_{t})$ is found by summing over the probabilities of each grasp for that goal:

\begin{equation}
\label{overall_prob}
p(\phi _{t} | \xi_{0\rightarrow t}) = \sum_{x \in C_{\phi}} p(x_{t} | \xi_{0\rightarrow t})
\end{equation}



\subsection{Determining State Transition Probabilities}
In this model, a transition probability between members of the same class versus different classes represents the probability of switching intended grasps on an object versus switching to a particular grasp of a new goal, respectively. In order to specify these probabilities from one timestep to the next, we use two parameters: $T_{grasp}$ is the probability of switching intended grasps on the same goal, and $T_{goal}$ is the probability of switching goals (where we constrain the parameter choices such that $T_{grasp} + T_{goal} \leq 1$). Then, we can calculate the transition probabilities going out of a state $x$ using the following equations:

\begin{equation}
p(x'_{t+1} | x_{t})=
\begin{cases}
\frac{T_{grasp}}{|C_{\phi}|-1} \text{, where } x,x' \in C_{\phi} \text{ and } x \neq x' \\[10pt]
\frac{T_{goal}}{|\Phi|-1}\cdot \frac{1}{|C_{\phi '}|} \text{, where } x' \in C_{\phi '}, x \not\in C_{\phi '} \\[10pt]
1 - (T_{grasp} + T_{goal}) \text{, where } x = x'
\end{cases}
\end{equation}

\subsection{Determining Observation Probabilities}
In order to perform filtering in this HMM, we need to specify the probability distribution over user actions for a given hidden state. That is, we need a way to calculate $p(u^{H}_{t} | x_{t})$, where $u^{H}_{t}$ is the user's action at timestep $t$. 

Each grasp $x$ consists of a particular position and orientation of the robot's end effector. Let the state of the robot's end effector at timestep $t$ be given by $s_{t}$. Then, we define a reward function $R_x (s, u)$ that estimates the value of choosing action $u$ in state $s$ while attempting to reach $x$. 

The implementation of this reward function can be designed to incorporate desired characteristics as our framework is agnostic to the specific implementation choice. We use a reward function defined by the anticipated difference in distance to the intended grasp after taking action $u$:
\begin{equation}
R_x (s, u) = dist(s, x) - dist(\tau (s, u), x)
\end{equation}
where $\tau (s, u)$ produces the resulting end effector state from applying action $u$ in state $s$.

We use a Boltzmann distribution to model the operator as exponentially more likely to choose actions with increasing reward, following prior work in Inverse Reinforcement Learning \cite{ramachandran2007bayesian} and assistive teleoperation \cite{jain2018recursive}:
\begin{equation}
p(u^H_{t} | x_{t}) = \frac{exp(R_{x_t} (s_{t}, u^H_{t}))}{\sum_{u \in U} exp(R_{x_t} (s_{t}, u))}
\end{equation}

\begin{figure}
\includegraphics[width=\columnwidth]{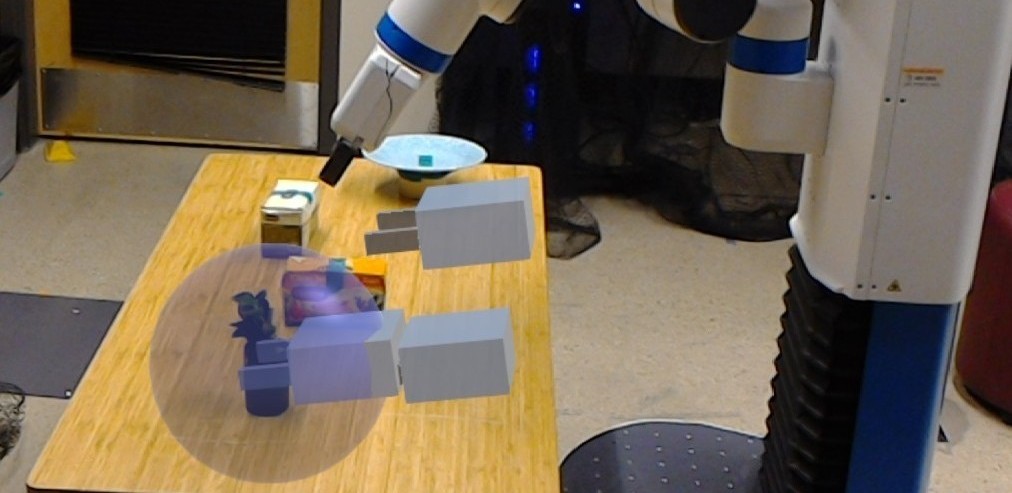}
\caption{A sphere around the target of assistance is intended to make the assistive controller's intent legible while keypoints of the assistive controller's intended trajectory are intended to make the system's future movement more predictable.}
\label{holograms_visualization}
\end{figure}

\subsection{Belief State Updating and Action Selection}

The use of an HMM allows us to perform iterative updating of the probability assigned to each particular grasp using the forward algorithm:
\begin{equation}
p(x_{t}|\xi_{0\rightarrow t}) = p(u^H_{t} | x_{t}) \sum_{x'}p(x'_{t-1}|\xi_{0\rightarrow t-1})p(x_{t}|x'_{t-1})
\end{equation}

Combining this with Eq. \ref{overall_prob} produces an updated estimate of the probability of each goal after each user action. 

In our system, if any goal is assigned a probability greater than $50\%$, the assistive controller selects the action $u_{r}$ that minimizes the expected distance toward an expert-demonstrated grasp of this target goal. In practice, these grasps could be generated by a planner using a robot's onboard perception system. The expert-demonstrated grasps we use are multi-point trajectories consisting of two to three keypoints that demonstrate a safe, reliable way to grasp this particular goal.

\subsection{Blending of Assistive Controller and Human Commands}
The final command sent to the robot is determined through the linear blending:

$$
u^{*} = \alpha u_{r} + (1-\alpha) u_{h}
$$

where $u_{r}$ is the command issued by the assistive controller, $u_{h}$ is the command issued by the user, and $\alpha$ is an \textit{arbitration} factor. This strategy of linear blending has an extensive history of usage in shared control \cite{dragan2013policy}.

Determining this arbitration factor impacts the final behavior significantly by setting the weighting between human and assistive controller. Another impact of this blending scheme is an effect on the robot's speed; if the assistive controller and user have opposite commands or if the user inputs no command, the robot's final speed will be slower in systems where $\alpha$ is not sufficiently close to 0 or 1. Consequently, we recognize that the effects of AR visualizations may differ based on the human-controller interactions produced by different arbitration factor values. We use multiple systems with differing values of $\alpha$ in this research in order to draw conclusions about the impact of our visualizations that are applicable to a range of shared control systems.

\subsection{Visualizations of Assistive Intent}
The visualizations that our system generates are focused on two goals: (1) improving legibility of the assistive controller's intent and (2) enhancing predictability of the shared controller's future movemement. While there are many possible ways to design visualizations for these two goals, for this study, we chose two visualizations that explicitly convey this information while requiring minimal user interpretation.

In order to improve legibility, our system displays a sphere around the inferred goal. This makes the assistive controller's inference immediately legible by explicitly calling out its goal of assistance. This sphere appears any time a goal hits $50\%$ probability in our goal inference system, centered on the newly inferred assistance target. Fig. \ref{holograms_visualization} shows an example of this sphere around the houseplant object.

To make the shared control system's movement predictable, we use virtual images of the robot's end effector that trace out the remaining keypoints in the expert-demonstrated grasp of this object. These visualizations directly illustrate the path that the assistive controller will attempt to follow. Fig. \ref{holograms_visualization} also contains an example of this type of visualization.

\section{User Study}

We collect data to test our hypotheses through a mixed-design 3 (visualization [between participants]: none, goal only, goal $+$ trajectory) $\times$ 3 (controller [within participants]: low assistance, medium assistance, high assistance) user study with $27$ participants (19 male, 8 female). In the study, participants interact with several shared control systems while operating a Fetch robot to grasp objects on a nearby table. Throughout this process, participants are shown one of three types of AR visualizations. The independent variable in which we are primarily interested is the visualization shown to participants, consisting of three conditions:

\subsubsection{No Visualization}
The participants are not shown any AR visualizations. This is our baseline condition representing how shared control systems work in practice today.

\subsubsection{Goal Only Visualization}
The participants are shown the goal sphere visualization whenever a goal is inferred.

\subsubsection{Goal + Trajectory Visualization}
The participants are shown both the goal sphere visualization and the virtual images of the assistive controller's intended path.

Our second independent variable represents the shared control systems with which the user interacts. The three shared controllers use the system described in Section \ref{System_Description} with three different $\alpha$-levels: $\alpha = 0.25$, $\alpha = 0.5$, $\alpha = 0.99$. We use $T_{grasp} = .01$ and $T_{goal} = 0$ to allow for transitions between target grasps while assuming a fixed goal. Participants use an $\alpha = 0$ (pure teleoperation) system first in order to establish a baseline level of control and subjective judgement of the robotic system, followed by the other three controllers in randomized order to mitigate any potential transfer effects (e.g., learning, fatigue).

\begin{figure*}
\centering
\includegraphics[width=\textwidth]{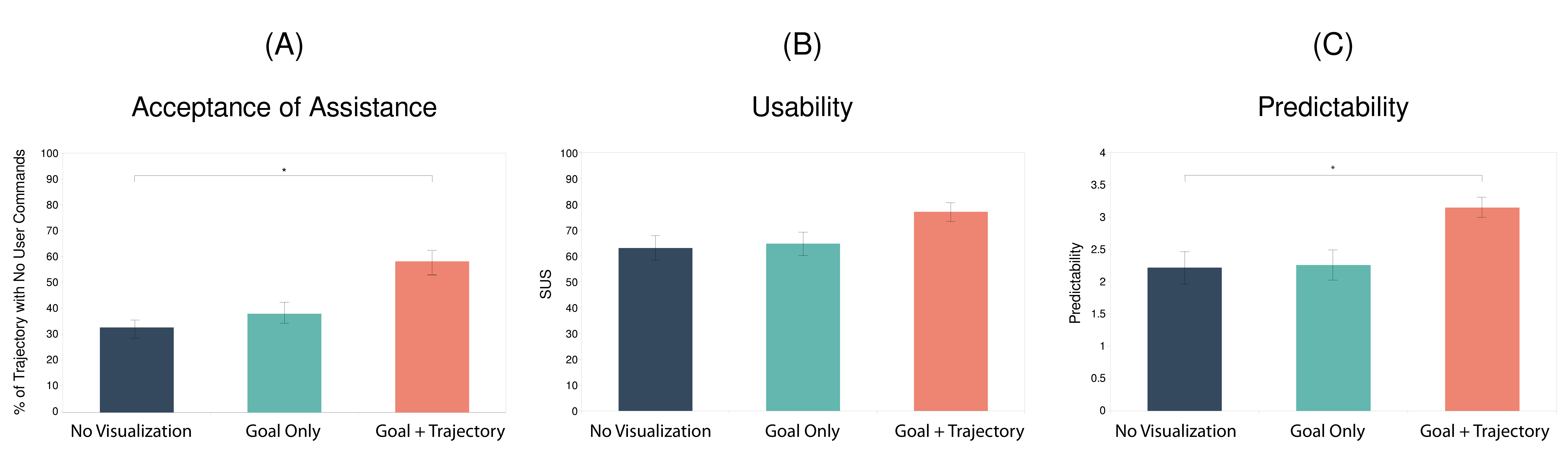}
\caption{(A) Users refrain from inputting control commands for a significantly larger percentage of the trajectory when shown the Goal + Trajectory visualizations. (B) Although there is an increasing average SUS score with more visualizations, we find no statistically significant effect of visualizations on perceived usability. (C) Our results shows a significant effect of visualization on user rating of the shared control system's predictability. This improvement is only found in the Goal $+$ Trajectory condition over the No Visualization baseline, with no significant difference between the baseline and the Goal Only visualization. In each of these subgraphs, $*$ signifies a results with a significance level of $p < 0.05$.}
\label{results_graph}
\end{figure*}

\subsection{Experiment Setup}

Participants operate the robot with a PlayStation controller using a modal system consisting of two control modes (position and angular control). This type of modal control is a common paradigm for teleoperation of high-DOF systems for which direct control over all joints at once may be impractical \cite{herlant2016assistive, gopinath2017mode}. Before the experiment begins, each participant is given a tutorial on how to use the controller, followed by a $2$ minute practice period. During the experiment, all participants wear a Microsoft HoloLens --- a popular AR head-mounted display (ARHMD) that we use to display the relevant visualizations for each participant's condition.

For each shared controller, participants operate the robot to grasp the four objects on the table. Before each grasp, participants are shown a letter in AR to instruct them which object to grasp next (this also provides a justification to participants in the No Visualization condition for the ARHMD). Participants are allowed four failed grasps on an object before it is skipped. Figs. \ref{teaser_visualization} and \ref{holograms_visualization} show the experimental setup consisting of four objects spread out on the table. 

The experiment consists of four rounds of grasping (once using pure teleoperation and once for each of the assistive controllers), each round including a single grasp of each of the four objects on the table. Each round of grasping is followed by completion of surveys on the shared control system just used. The entire experiment thus consists of performing $16$ successful grasps and $4$ rounds of surveys, taking approximately $30$ minutes per participant.

\subsection{Measures}

In order to evaluate our hypotheses, we assess both objective and subjective metrics that provide insight into the joint behavior of the human and assistive controllers.

The objective metrics that we collect data on are \textit{completion effort}: the total number of user actions required per successful grasp, and \textit{acceptance of assistance}: the percentage of the trajectories from each successful grasp during which the user does not actively input controls. As a manipulation check, we compare \textit{completion effort} across controller types to test if our shared control systems improve user efficiency. We use \textit{acceptance of assistance} to evaluate the willingness of users to relinquish control to the assistive system (\textbf{H1}).

We administer a survey after participants interact with each controller to measure subjective participant experience. We use the System Usability Scale (SUS) in order to utilize a validated metric for assessing the perceived \textit{usability} of a technology interface (\textbf{H2}). Additionally, after participants use each controller with an $\alpha > 0$, we record responses on a 2-item scale (Cronbach's $\alpha$ = 0.869), which consists of 5-point Likert-style questionnaire items ("I understood why the robot moved the way it did", "I found the robot's movement predictable"), to evaluate subjective perceptions of the shared control system's \textit{predictability} (\textbf{H3}).

\section{Analysis and Results}
After $27$ participants completed our study, we observed that only one participant failed the limit of four times on a particular grasp (No Visualization condition, $\alpha=0.99$ controller). Generally, we conduct a one-way analysis of variance (ANOVA) and find no significant difference in the number of grasp failures between visualization conditions F(2,24) = 1.09, p = 0.353. We receive no successful trajectory data from this failed grasp as well as one other grasp for which our recording crashed, leaving us with data from $322$ successful grasps between the 3 shared controllers.

For each of the measures discussed previously, we conduct a mixed-design ANOVA, treating visualization type as the fixed effect and controller type as the within-subjects variable. If the ANOVA demonstrates a significant effect of the independent variable of interest on a measure, we conduct a \textit{post hoc} Dunnett's test to identify which conditions demonstrate significant differences from the baseline.

\subsection{Objective Results}

We find that controller type has a significant effect on \textit{completion effort} F(3,72) = 13.05, p $<$ 0.0001. A \textit{post hoc} Dunnett's test against the $\alpha = 0$ baseline (M = 88.77, SD = 35.85) shows significant improvements for $\alpha = 0.5$ (M = 65.07, SD = 54.28, p = 0.014) and $\alpha = 0.99$ (M = 43.62, SD = 45.17, p $<$ 0.0001) controllers, but none for the $\alpha = 0.25$ controller (M = 85.48, SD = 38.70, p = 0.959). This confirms our manipulation checks that our assistive controllers improve user efficiency at higher levels of $\alpha$ irrespective of visualization condition. 

While the previous measure assesses overall user control efficiency, \textbf{H1} is concerned with how the human user cooperates with the assistive controller. Our results demonstrate that visualization has a significant effect on \textit{acceptance of assistance} F(2,24) = 5.10, p = 0.014. A post-hoc Dunnett's test against the No Visualization (M = 32.73, SD = 18.32) baseline demonstrates that the Goal $+$ Trajectory condition (M = 58.19, SD = 22.64) significantly improved user acceptance of assistance (p = 0.010) while Goal Only (M = 38.77, SD = 19.59) does not (p = 0.694). These results show support for \textbf{H1} when using both visualizations to increase goal legibility and trajectory predictability while not finding any such support with visualizations designed to increase legibility alone. Fig. \ref{results_graph} (A) illustrates these results.

\subsection{Subjective Results}

We use the validated SUS scale to measure subjective ratings of the shared control systems' \textit{usability}. We do not find a significant effect of visualization on this measure F(2, 24) = 1.84, p = 0.181 and thus do not find support for \textbf{H2}. The mean SUS scores for No Visualization (M = 63.06, SD = 24.47), Goal Only (M = 64.72, SD = 23.31), and Goal $+$ Trajectory (M = 77.04, SD = 18.89) are shown in Fig. \ref{results_graph} (B).

The final measure we assess is the user's subjective rating of the shared control system's \textit{predictability}. Our results reveal a significant effect of visualization on this rating F(2, 24) = 4.25, p = 0.026. A post-hoc Dunnett's test against the No Visualization (M = 2.22, SD = 1.31) baseline condition again demonstrates that the Goal $+$ Trajectory (M = 3.15, SD = 0.806) visualization produces a significant improvement (p = 0.031) while Goal Only (M = 2.26, SD = 1.18) visualization does not (p = 0.992). Consequently, we find support for \textbf{H3} when using both types of visualizations while again not finding any such support for legibility visualizations alone. These results are displayed in Fig. \ref{results_graph} (C).

\subsection{Discussion}
Our results demonstrate that users find shared control systems more predictable when given visualizations of the assistive controller's inferred goal and intended trajectory. In these cases, participants are more willing to hand over control to the assistive system, creating trajectories with less direct human control, although they do not rate these systems subjectively higher in usability. Thus, we find evidence for \textbf{H1} and \textbf{H3} specifically when using both types of visualizations, but we do not find support for \textbf{H2}. These findings demonstrate the potential for visualizations of assistive controller intent to improve cooperation between humans and assistive controllers in shared control systems while highlighting the complexities of improving user preference for these systems.

We find no evidence supporting our hypotheses when using only visualizations intended to improve assistive controller legibility. This may show that in order to obtain the benefits of users properly understanding a shared control system, users need to understand both the target and the intended trajectory of assistance. Although further work is needed to confirm this trend, our results suggest that legibility alone may not improve user cooperation with an assistive controller unless it is also accompanied by user-predictable behavior.

\section{Future Work}
Our results provide insight into some promising areas for further research. For instance, future work could investigate how our visualizations affect the actual actions taken by the human user. Shared control systems can influence users during cooperative control \cite{flemisch2019joining}, raising the possibility that visualizing the assistive controller's intended trajectory could affect the user's intended trajectory. Future work investigating whether or not this takes place would help better explain the mechanism of cooperation in systems utilizing our visualizations.

Since our results demonstrate that visualizations can provide benefits to shared control, this creates an open question about how to best implement these visualization for different types of shared control. We create and evaluate two representative types of visualizations, but a more thorough exploration of this design could be enlightening. Furthermore, future work could examine the combination of these visualizations with other common feedback systems for shared control such as haptic feedback. Finally, as our visualizations that are intended to improve predictability also increase legibility, we leave the effect of predictability visualizations that do not improve legibility as a question for future work.

\section{Conclusion}
Through shared control, robot autonomy can be used to empower a human user of a robotic system without removing the user from the control loop. While much work in shared control is focused on improving the assistive controller's understanding of the user's intent (e.g. goal inference), this work introduces and evaluates a way to improve the user's understanding of the assistive controller's intent. Our results demonstrate the potential for positive impact from these visualizations as well as the importance of sufficient information communication for this impact to be seen. This also has implications for shared control beyond the design of AR visualizations: the legibility of an assistive controller, without predictability, may not be sufficient for improving acceptance of shared control. We look forward to seeing these types of visualizations further explored in future work on improving shared control of robotic system.

\section{Acknowledgements}
This work was supported by a grant from the NSF under award 1764092.

\bibliographystyle{IEEEtran}
\bibliography{refs}

\end{document}